\documentclass[10pt,twocolumn,letterpaper]{article}

\usepackage{wacv}
\usepackage{times}
\usepackage{epsfig}
\usepackage{graphicx}
\usepackage{amsmath}
\usepackage{amssymb}
\usepackage[accsupp]{axessibility}
\usepackage{color}
\usepackage{caption}
\usepackage{booktabs} 
\usepackage{array}
\usepackage{relsize}
\usepackage[inline]{enumitem} 
\usepackage{color}
\usepackage{varwidth}
\usepackage{multirow}
\usepackage{slashbox}

\usepackage{lipsum}

\newcommand\blfootnote[1]{%
  \begingroup
  \renewcommand\thefootnote{}\footnote{#1}%
  \addtocounter{footnote}{-1}%
  \endgroup
}

\def\wacvPaperID{968} 

\wacvfinalcopy 

\ifwacvfinal
\fi

\usepackage[pagebackref=true,breaklinks=true,colorlinks,bookmarks=false]{hyperref}

\newcolumntype{?}{!{\vrule width 1pt}}

\begin{document}

\title{Enhanced Correlation Matching based Video Frame Interpolation}

\author{$\text{Sungho Lee}^{*}$
\qquad
$\text{Narae Choi}^{*}$
\qquad
Woong Il Choi\\
Samsung Research, South Korea\\
}

\maketitle
\ifwacvfinal
\thispagestyle{empty}
\fi
\begin{abstract}
We propose a novel DNN based framework called the Enhanced Correlation Matching based Video Frame Interpolation Network to support high resolution like 4K, which has a large scale of motion and occlusion. 
Considering the extensibility of the network model according to resolution, the proposed scheme employs the recurrent pyramid architecture that shares the parameters among each pyramid layer for the optical flow estimation. 
In the proposed flow estimation, the optical flows are recursively refined by tracing the location with maximum correlation. 
The forward warping based correlation matching enables to improve the accuracy of flow update by excluding incorrectly warped features around the occlusion area.
Based on the final bi-directional flows, the intermediate frame at arbitrary temporal position is synthesized using the warping and blending network and it is further improved by refinement network.
Experiment results demonstrate that the proposed scheme outperforms the previous works at 4K video data and low-resolution benchmark datasets as well in terms of objective and subjective quality with the smallest number of model parameters.
\end{abstract}
\blfootnote{* indicates equal contribution.}
\section{Introduction}

Video frame interpolation (VFI) is a task to generate intermediate frames between consecutive frames and its main application is to increase frame-rate~\cite{meyer2015phase} or to represent slow-motion video~\cite{jiang2018super}.
It is also applicable to synthesize novel views of 3D contents~\cite{flynn2016deepstereo}. Advanced deep learning model based approaches have been introduced to achieve a better quality of the interpolated frames by discovering knowledge from large-scale diverse video scenes.

However, those approaches have a limitation on the range of motion coverage because the network only explores the motion of pixels within the receptive field of the CNN model that is pre-defined by their kernel size. 
As the most of capturing and display devices already support 4K resolution, the support of higher resolution is an inevitable feature of the practical application of VFI. 
The limitation of handling large motion will become more pronounced as the image resolution increases and will be a major problem of practical use.

We can simply consider increasing the receptive field of a model by increasing kernel size or depth of network which attend utilizing more parameters to overcome the limitation of motion range. However, it is an inefficient solution because the memory size and computational complexity will be increased accordingly. 
Instead of expanding network kernel, the pyramid based architecture~\cite{sun2018pwc, teed2020raft} has been proposed which progressively updates optical flow from coarse-scale to fine. Even though the network of each layer has a limited receptive field, the pyramid structure can extend the motion coverage by adding the number of pyramid layers. Recently, a recurrent model based pyramid network has been proposed to avoid the model parameters increasing according to the number of layers~\cite{sim2021xvfi, zhang2020flexible}. It basically re-uses the fixed model parameters in each layer, and the optical flows are iteratively refined from residual flows using the same network.

We propose an enhanced correlation matching based video frame interpolation network (ECMNet) to support high-resolution video with the lightweight model. The proposed network consists of three sub-networks; 
The first one is the bi-directional optical flow estimation between two consecutive input frames. Inspired by recurrent pyramid structure, this module adopts the shared pyramid structure to extend the receptive field efficiently. Also, we present an enhanced correlation matching algorithm to improve the accuracy of flow estimation.
The next part is a frame synthesis module to generate intermediate flows and frame at an arbitrary temporal position. The last is a refinement module to improve the interpolated frame. Unlike the first module, the last two modules operate once with the original scale of input frames. This leads to computational efficiency and reduction of network parameters.
In summary, the major contributions of our paper are as follows:
\begin{itemize}
    \item We propose an enhanced correlation computation method based on the forward warped feature for flow estimation.
    \item At each pyramid layer, flows are gradually updated using the highest feature correlation based on a novel feature matching algorithm.
    \item The proposed network achieves the state-of-the-art performance with the fewest number of parameters at 4K dataset compared to the existing VFI methods.
\end{itemize}

\begin{figure*}
\centering
\includegraphics[width=1.0\linewidth]{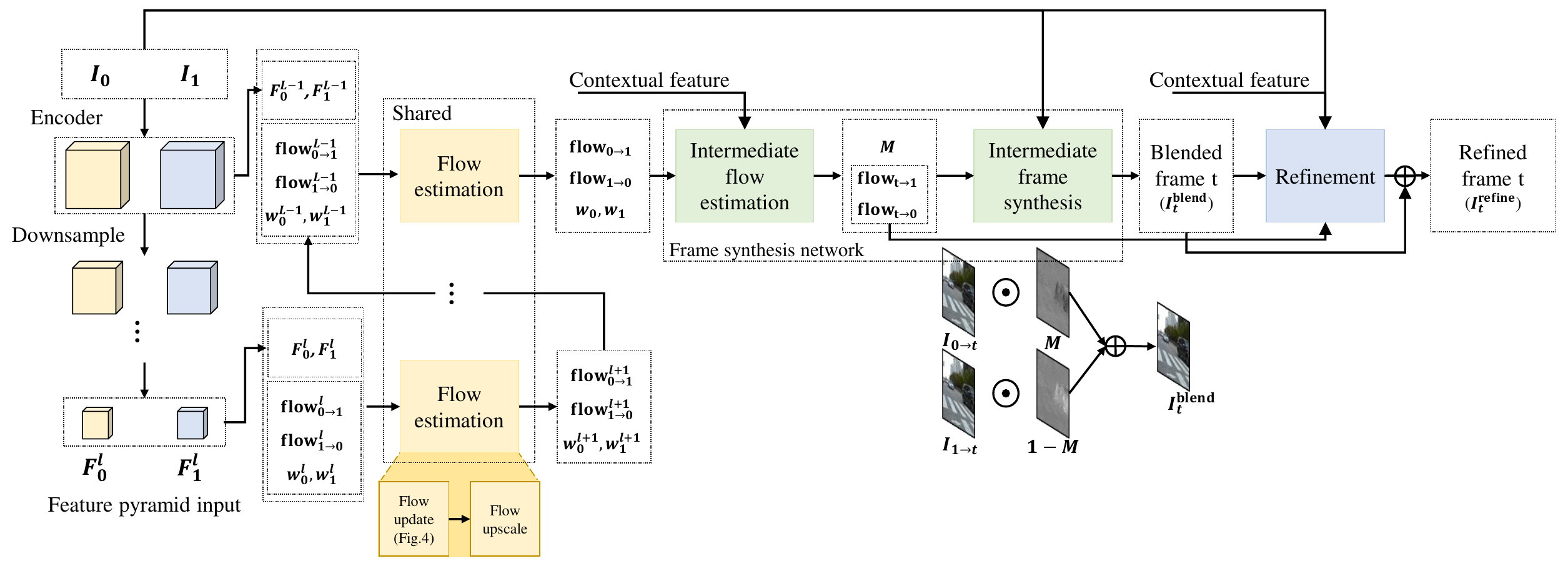}
\caption{Overview of the network. 
Our framework consists of three sub-networks; Bi-directional flow estimation network, frame synthesis network, and refinement network. 
Shared pyramid network for flow estimation is denoted to yellow boxes. 
Green boxes indicates frame synthesis network which computes intermediate flow and initial intermediate frame for the arbitrary temporal position $t \in (0,1)$. 
Finally, refinement network corrects the intermediate frame using estimated residual. 
Note that ${\boldsymbol{w}_n}$~($n = 0$ or $1$) is the learnable importance weight in $\boldsymbol{F}_n$ coordinate to resolve the forward warping issues, where multiple pixels from $\boldsymbol{F}_n$ map to the same target pixels in $\boldsymbol{F}_{1-n}$~\cite{niklaus2020softmax}, and learnable blending mask $\boldsymbol{M}$ is the confidence ratio of the warped image $\boldsymbol{I}_{0 \rightarrow t}$. 
}
\label{fig:overview}
\end{figure*}
\section{Related works}

\subsection{Video frame interpolation}
In this section, we focus on recently proposed learning based algorithms. It can be categorized as kernel based and flow based approaches. Long \etal\cite{long2016learning} firstly introduced a generic CNN model to directly generate an intermediate frame. Since those generic models are not enough to exploit various motion characteristics of natural video, the result images are damaged by blur artifacts.
It is enhanced by AdaConv and SepConv methods proposed by Niklaus \etal\cite{niklaus2017video}, which generate interpolated pixels by locally convolving the input frames.
They formulated pixel interpolation as convolution over pixel patches instead of relying on optical flow, and it is able to synthesize pixels from relatively large neighborhoods than previous work. AdaCoF~\cite{lee2020adacof} proposed to apply a deformable convolution kernel and CAIN~\cite{choi2020channel} proposed to utilize a feature reshaping operation, PixelShuffle with channel attention to synthesis image. These kernel based approaches tend to have a relatively simpler design of network model, while those have a limitation on the range of motion coverage because the pixels that are located out of range of the receptive field could not participate in interpolation. If it is tried to increase the coverage of large motion in high resolution, it requires a high memory footprint and heavy computational complexity.

The second approach is to estimate the motion flow of each pixel between consecutive frames and synthesize an image from the corresponding pixels guided by the estimated flow. 
Liu \etal\cite{liu2017video} proposed the Deep Voxel Flow (DVF) to perform trilinear interpolation using the spatio-temporal CNN model.
Bao \etal\cite{jiang2018super} proposed Super-Slomo which combines U-Net architecture of DVF and kernel based image synthesis between warped images.
The main architecture in Super-Slomo~\cite{jiang2018super} is the cascade model of bi-directional flow estimation between input frames and image synthesis to blend two warped images guided by estimated bi-directional flows and it becomes a standard framework of flow based frame interpolation methods.
However, those flow estimation based on U-Net structure still has a limitation on handling large motion. Instead of U-Net~\cite{ronneberger2015u}, it has been proposed to employ an advanced flow estimation model like PWC-Net~\cite{sun2018pwc} applied iterative refinement using coarse-to-fine pyramid structure.

For the image synthesis part, a key issue is how to properly handle the occlusion area during blending warped pixels.
MEMC-Net~\cite{bao2019memc} implicitly handles occlusion by estimating occlusion masks from features of input frames, while DAIN~\cite{bao2019depth} explicitly detects occlusion from estimated depth information.
From depth, it is determined whether the object is foreground or background and it is possible to select pixels from the foreground object in the area where both objects are overlapped.
The performance should depend on the accuracy of depth and it is still challenging to obtain accurate depth from a single image.

While most flow based approaches perform backward warping for image synthesis, it is proposed in Softmax splatting~\cite{niklaus2020softmax} to perform forward warping by solving the ambiguity on overlapping the pixels from different flows to the same pixel.

\subsection{Flow Estimation Network for Large Motion}
From traditional approaches of an optical flow estimation, multi-resolution based approaches were already introduced which refine the resolution of flows from coarse-scale to fine in order to avoid drastically increased computational complexity of full search~\cite{bruhn2005lucas}. 
In contrast to traditional coarse-to-fine approaches, PWC-Net~\cite{sun2018pwc} utilizes a learnable feature pyramid instead of image pyramid to overcome the variation of lighting changes or shadow. 
Since it has been the state-of-the-art method until RAFT~\cite{teed2020raft} is introduced in 2020, PWC-Net has been mainly adopted in flow based VFI scheme~\cite{bao2019depth, niklaus2020softmax, xu2019quadratic}. 
RAFT~\cite{teed2020raft} pointed out the drawbacks of the coarse-to-fine approach; the difficulty of recovering estimation error at a coarse resolution and the tendency to miss the small fast-moving object.
It proposes a multi-scale correlation based estimation approach to overcome those issues. 
However, those approaches still have limitations in two aspects; one is that network size increases in proportion to the number of pyramid layers and the other is that it is required to re-train the network if the target resolution is different to the resolution of trained images. 
To overcome those issues, RRPN~\cite{zhang2020flexible} proposed the recurrent pyramid model to share the same weights and base network among different pyramid layers. 
Compared to RRPN~\cite{zhang2020flexible}, XVFI-Net~\cite{sim2021xvfi} proposes more generalized model to synthesis frames at the arbitrary temporal position because RRPN was only applicable to synthesize the center frame.

The network architecture of the proposed {ECMNet} is designed based on the recurrent pyramid approach as well. 
Compared to RRPN and XVFI-Net~\cite{sim2021xvfi,zhang2020flexible}, our {ECMNet} applied correlation matching based flow estimation instead of directly deriving flows from U-Net.
Compared to PWC-Net and RAFT~\cite{sun2018pwc,teed2020raft}, the correlation cost is constructed by comparing forward warped feature instead of backward warped one. 
Besides, while the existing methods~\cite{sun2018pwc,teed2020raft} derive the flows by the network with cost volume as an input, the proposed method determines the optical flows directly by checking the maximum correlation value.

\section{Proposed algorithm}
The overview of our framework is shown in figure~\ref{fig:overview}. 
Our network takes consecutive images $\boldsymbol{I}_{0}$ and $\boldsymbol{I}_{1}$, and the desired temporal position $t \in (0,1)$ as inputs to synthesize the intermediate frame $\boldsymbol{I}_{t}$. 

In the flow estimation module, optical flows between input frames are estimated by comparing two encoded features.
The intermediate flow estimation network then computes optical flows from the temporal position $t$ to each input. 
Given the input images and intermediate flows, warping and blending are performed to synthesis the output frame at time $t$. 
Finally, the output image of the frame synthesis network is further improved at the following refinement network.
The flow estimation module adopts a pyramid structure to compute the optical flow for input images from low to original resolution.
After flow estimation, intermediate frame synthesis and refinement are  performed only with the original scale of input images. 
This structure enables frame synthesis at any temporal position with estimated bi-directional optical flows.
The details of the network architecture are described in the supplementary material.

\begin{figure}
\centering
\includegraphics[width=1.0\linewidth]{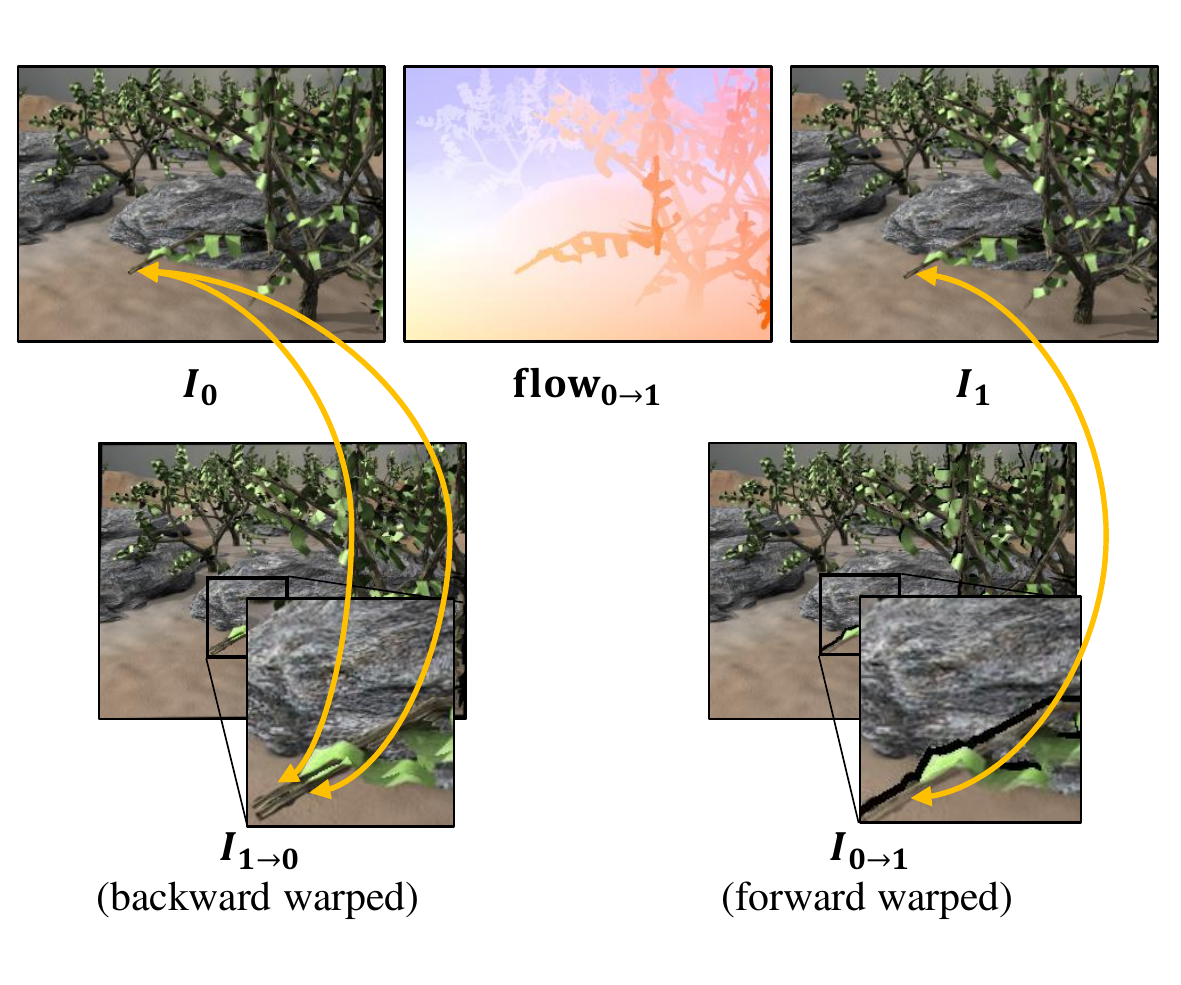}
\caption{Backward and forward warping examples~\cite{baker2011database}. As shown in $\boldsymbol{I}_{1 \rightarrow 0}$, backward warping creates duplication where the flow fail to map pixels to the shifted location such as occlusion region. The duplication makes difficulties to compare the similarity between the warped image and original image. On the other hand, forward warping doesn't cause confusion because it does not create unknown parts.}
\label{fig:warp_example}
\end{figure}

\subsection{Bi-directional optical flow estimation}
\label{section:flow}
Optical flow estimation for frame interpolation can be categorized as U-Net based and correlation based methods. 
U-Net based methods~\cite{sim2021xvfi, zhang2020flexible} have a relatively simple structure but are more difficult to explicitly compare input features than correlation based methods~\cite{sun2018pwc, teed2020raft, xu2017accurate}.
The correlation based one utilizes the feature correlation as an input of the flow estimation network to guide the warped feature matches to the original feature. 
However, these algorithms are difficult to ensure that the output flow is corrected to a position with the highest correlation. 
Therefore, we propose the flow correction with the highest feature correlation based on novel feature matching algorithms in order to estimate bi-directional flows. 
To estimate more accurate flows, forward warping is performed to compute feature correlations. 
Inspired by softmax splatting approach~\cite{niklaus2020softmax}, the learnable importance weights $\boldsymbol{w}^{l+1}_{0}$ and $\boldsymbol{w}^{l+1}_{1}$ for blending forward warped pixels are derived at each pyramid layer.

The bi-directional flow estimation network consists of an encoder, feature downsampling network, a flow update module, and a flow upscaling network.

\paragraph{Encoder and feature downsampling network}  
Encoder network extracts the features $\boldsymbol{{F}}^{l}_{0}$ and $\boldsymbol{{F}}^{l}_{1}$ from each input images $\boldsymbol{I}_{0}$ and $\boldsymbol{I}_{1}$. 
We denote the pyramid level as $\boldsymbol{l}$. 
To generate the input feature for each pyramid layer, the strided convolution network is used to downsample the features by a factor of 2. 

\begin{figure}
\centering
\includegraphics[width=1.0\linewidth]{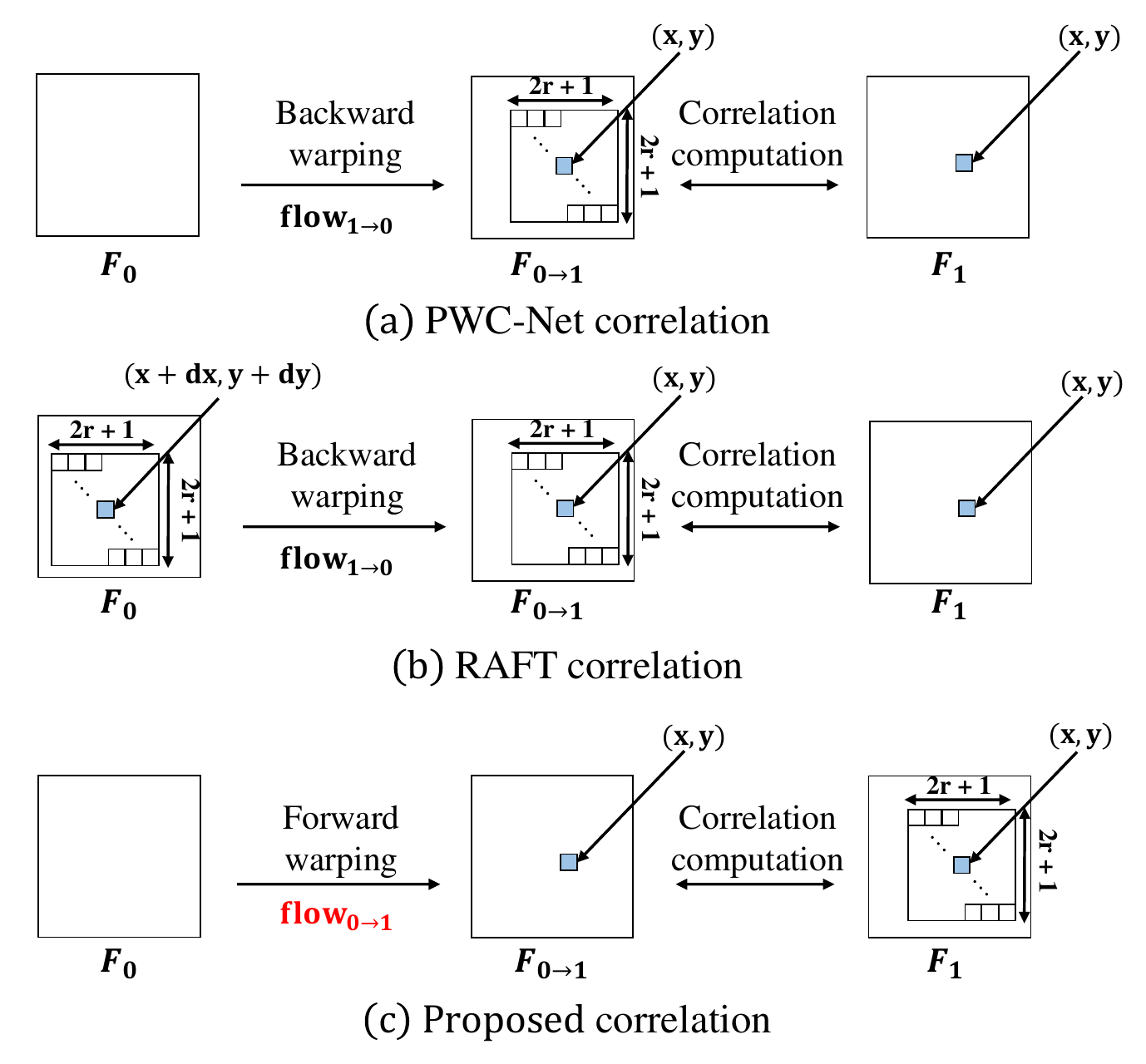}
\caption{Comparison of correlation matching method between existing and the proposed one. Note that the local grid $(2r+1)\times(2r+1)$ in the features indicates the search range to find motion flows through correlation comparison.  (a) PWC-Net~\cite{sun2018pwc} performs backward warping to compute $\boldsymbol{F}_{0\rightarrow1}$ and correlation matching is followed. 
(b) RAFT~\cite{teed2020raft} extracts the candidate from the corresponding location of $(x,y)$ in the $\boldsymbol{F}_{0}$, candidates are warped using $\boldsymbol{F}_{1\rightarrow0}(x,y)$ to align the spatial coordinate. (c) Our model performs forward warping to compute $\boldsymbol{F}_{0\rightarrow1}$, and the correlation is derived with the candidates extracted from $\boldsymbol{F}_{1}$. }
\label{fig:correlation}
\end{figure}

\begin{figure}
\centering
\includegraphics[width=1.0\linewidth]{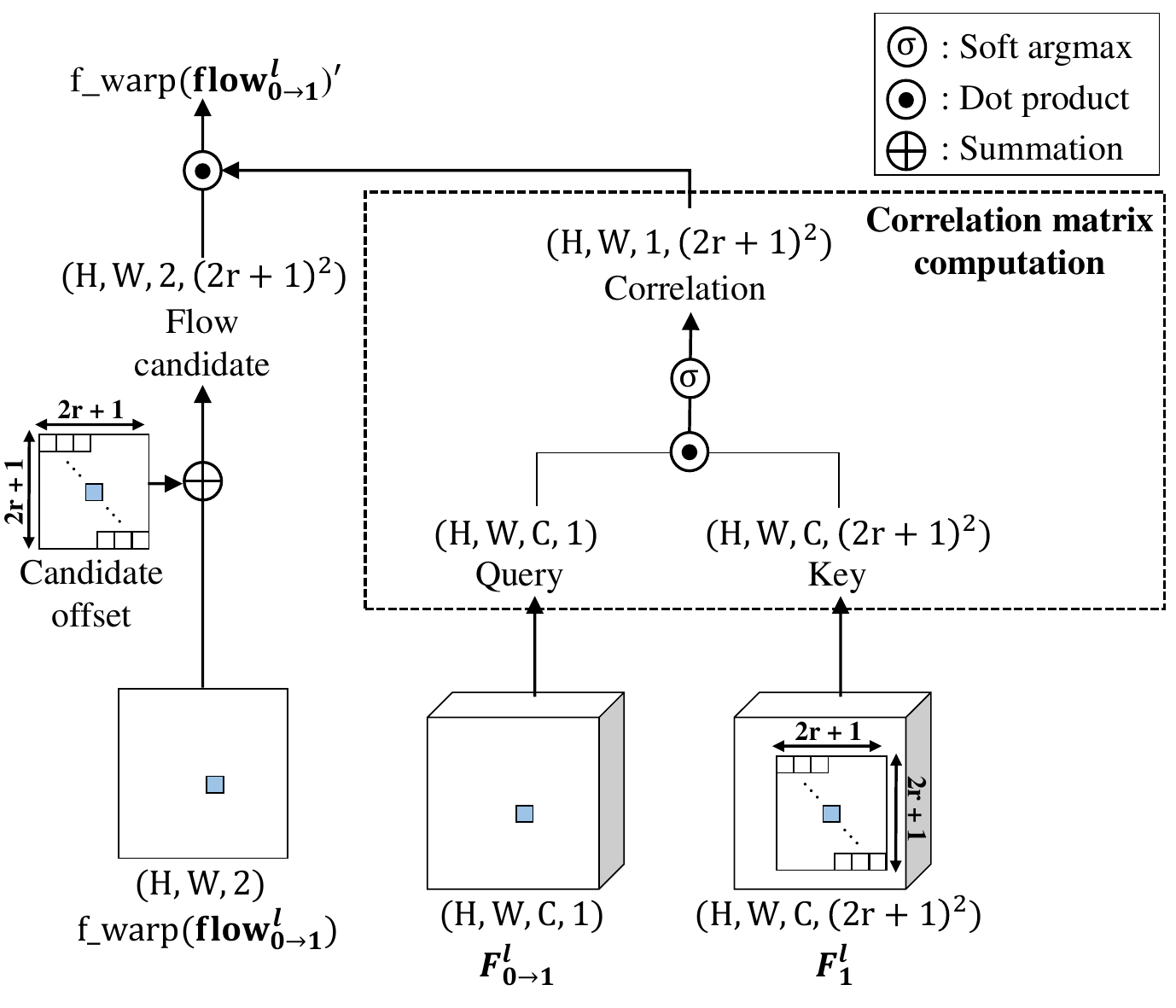}
\caption{Flow update process. Correlations of candidates are calculated with dot product according to the feature channel. Based on the soft-argmax operation on the correlation matrix, it is possible to determine the candidate having the highest similarity. After that, flow is updated with the selected candidate.
}
\label{fig:flowUpdate}
\end{figure}

\label{section:flowupdatemodule}
\paragraph{Flow update module}
The bi-directional flows are updated through each pyramid layer from the initial flows $\boldsymbol{\mathrm{flow}}^{1}_{0 \rightarrow 1}$ and $\boldsymbol{\mathrm{flow}}^{1}_{1 \rightarrow 0}$ which are set to zero. 
The module computes similarities between the input features as the correlation.
The flows from the previous pyramid level are updated by the comparison of correlation values.
To align the coordinates of two features for deriving correlation, forward warping is applied in contrast to the existing methods~\cite{sun2018pwc, teed2020raft} which use backward warping.
Figure~\ref{fig:warp_example} illustrates backward warped image has duplicated region when it fails to map the corresponding pixels. 
These miss-aligned features make ambiguity in determining the optimal values of the correlation. 
Since the forward-warped feature does not include these miss-aligned features, it is possible to estimate more accurate flows.
Figure~\ref{fig:correlation} and~\ref{fig:flowUpdate} describe the correlation calculation and flow update process at a certain spatial location $(x,y)$.
The proposed correlation matching method shown in figure~\ref{fig:correlation}(c) calculates feature correlation between forward warped feature $\boldsymbol{{F}}_{0 \rightarrow 1}$ and the input feature $\boldsymbol{{F}}_{1}$. 
Since the location of warped feature $\boldsymbol{{F}}_{0 \rightarrow 1}$ is aligned to the feature $\boldsymbol{{F}}_{1}$, the correlation is computed between the feature $\boldsymbol{F}_{0 \rightarrow 1}$ at $(x,y)$ and the feature $\boldsymbol{F}_{1}$ at $(x', y')$, where $(x',y') \in \{x-r \leq x' \leq x+r, y-r \leq y' \leq y+r\}$ with search range $r$. 
The flows are corrected by selecting one having the highest correlation value from the candidates.

The details of the flow update process are shown in figure~\ref{fig:flowUpdate}. 
The module calculates the similarity inside the search range by dot product according to the feature channel.
This procedure is similar to derive the relevance between `query' and `key' in the attention mechanism~\cite{vaswani2017attention}. 
Applying soft-argmax operation enables the flow to be determined the highest correlation in a differential way. 
At each pyramid layer, the flow is gradually updated using the highest feature correlation within the limited search range. 
Since the updated flow is projected to the spatial coordination of $\boldsymbol{F}^{l}_{1}$, it should be re-warped to the original coordination. 
We omit the re-warping process of the updated flow in figure~\ref{fig:flowUpdate} for compact expression. 
The optical flow of the opposite direction is also estimated by the same process to obtain bi-directional optical flows.

\paragraph{Flow upscaling network}

\label{section:flowupscalingnetwork}
Flow upsampling is required to use the flows of the previous pyramid level estimated with lower resolution. 
Since simple bi-linear interpolation leads to degradation of flow details, we design the network to minimize the error that occurred in the upsampling process.
The network takes inputs of the flow update module outputs, the importance weight outputs of the previous level, and the features of pyramid levels.
The network consists of decoder network and two sub-networks to estimate the residuals and importance weights for the current pyramid level. 
The residuals contribute to improving the quality of the upsampled flows.

\subsection{Frame synthesis network}
Intermediate frame is derived using two input frames and estimated bi-directional flows ($\boldsymbol{\mathrm{flow}}_{0 \rightarrow 1}$ and $\boldsymbol{\mathrm{flow}}_{1 \rightarrow 0}$). 
Green boxes of figure~\ref{fig:overview} describe frame synthesis network which consists of intermediate flow estimation network and frame synthesis layer.
Frame synthesis network performs once at the original scale of input images, unlike flow estimation module in section~\ref{section:flowupscalingnetwork} adopts pyramidal structure. 
We consider if the bi-directional flows are already estimated with enough receptive field which can be managed with a shared pyramidal structure, frame synthesis does not need to repeat with multi-scaled input images. 
We firstly calculate the intermediate flows $\boldsymbol{\mathrm{flow}}_{t \rightarrow 0}$ and $\boldsymbol{\mathrm{flow}}_{t \rightarrow 1}$ and then warping and blending are performed to interpolate the target frame $\boldsymbol{{I}}_{t}$.

\paragraph{Feature extraction network}
Frame synthesis network uses contextual information to improve the details and to deal with occlusions. 
The contextual feature is generated using two kinds of features, one is extracted from the encoder of the flow estimation module, and the other is extracted from the $conv1$ layer of the pre-trained VGG16~\cite{simonyan2014very}. 
The final contextual feature presented in figure~\ref{fig:overview} is obtained through a convolution layer that has inputs of the aforementioned extracted features, two input images, and flows.

\paragraph{Intermediate flow estimation network}
We use backward warping to synthesis an intermediate frame to avoid hole generation.
Since backward warping needs a flow that is starting from the temporal target position $t$ to input images, we adopt the flow reversal approach~\cite{xu2019quadratic} to estimate a flow aligned to coordinate at $t$.
Since it is possible for flow reversal to generate holes in the reversed flow mostly around the occluded area due to forward warping, flow residuals are computed to compensate for each reversed flow.
Additional refinement for the derived flows, such as the weighted average of corresponding flows, is not performed to maintain a characteristic of each flow.

\begin{equation} \label{eq:flow_t_0}
\begin{aligned}
    \mathcal{\boldsymbol{\mathrm{flow}}}_{t \rightarrow 0} &= \text{rev}({{{t} \cdot \boldsymbol{\mathrm{flow}}}}_{0 \rightarrow 1}) + \Delta{\boldsymbol{\mathrm{flow}}}_{t \rightarrow 0},\\
    \mathcal{\boldsymbol{\mathrm{flow}}}_{t \rightarrow 1} &= \text{rev}({{(1-{t}) \cdot \boldsymbol{\mathrm{flow}}}}_{1 \rightarrow 0}) + \Delta{\boldsymbol{\mathrm{flow}}}_{t \rightarrow 1},
\end{aligned}
\end{equation}
where the function `rev' denotes the flow reversal, and $\boldsymbol{\Delta{\mathrm{flow}}}_{t \rightarrow 0}$ and $\boldsymbol{\Delta{\mathrm{flow}}}_{t \rightarrow 1}$ are the flow residuals.

To estimate the flow residuals, we use the U-Net architecture composed of three layers of down-convolution and three layers of up-convolution. 
The network takes the bi-directional flows and the contextual feature as inputs, and generates residuals of intermediate flows and blending mask $\boldsymbol{{M}}$.

\paragraph{Intermediate Frame Synthesis}  
We simply perform frame synthesis with warping and blending. 
In the warping process, two input images are backward warped to time ${t}$ using estimated intermediate flows. 
The warped images $\boldsymbol{{I}}_{0 \rightarrow t}$ and $\boldsymbol{{I}}_{1 \rightarrow t}$ usually contain artifacts especially the region which is unseen at the input images and appeared at the time ${t}$. 
To overcome the issue, we blend two warped images with the learnable blending mask~$\boldsymbol{M}$ as follows.
\begin{equation} \label{eq:frameinterpolation}
\begin{aligned}
    \mathcal{\boldsymbol{I}}^\text{blend}_{t} = ({{\boldsymbol{M} \odot \boldsymbol{I}}_{0 \rightarrow t}}) + ({{(1-\boldsymbol{M}) \odot \boldsymbol{I}}_{1 \rightarrow t}}),
\end{aligned}
\end{equation}
where $\odot$ is the pixel-wise multiplication operation, and blending mask $\boldsymbol{\mathrm{M}}$ denotes the confidence ratio of the warped image $\boldsymbol{I}_{0 \rightarrow t}$.
The confidence ratio has a value between 0 and 1.

\subsection{Refinement network}  
The refinement network enhances the interpolated output given the blended intermediate frame, contextual feature, input images, and intermediate flows.
Optical flow estimation can be failed in the occlusion, blurred, and brightness changed area. 
Most of the artifacts caused by flow estimation failure can be resolved by blending, but the artifacts may occur in the same area of $\boldsymbol{I}_{0 \rightarrow t}$ and $\boldsymbol{I}_{1 \rightarrow t}$.
To solve this case, we design the network adding the dilated convolutional block to grow the receptive field inspired by image inpainting algorithm~\cite{iizuka2017globally}. 
The network completes the occluded region and enforces the network to predict the residuals between the ground-truth and blended output.

\subsection{Loss Functions}
The proposed network illustrated in figure~\ref{fig:flowUpdate} is trained using the following two loss functions; One is to measure how close the blended intermediate frame~${\boldsymbol{I}^\mathrm{blend}_{t}}$ to the ground truth frame ${\boldsymbol{I}^{GT}_{t}}$, and the other is to compare the refined intermediate frame~${\boldsymbol{I}^\mathrm{refine}_{t}}$ with the frame ${\boldsymbol{I}^{GT}_{t}}$. 

\begin{equation} \label{eq:loss}
\begin{aligned}
    \mathcal{L}_\mathrm{blend} &= { \sum_{x} ||{\boldsymbol{I}^\mathrm{blend}_{t}(x) }-{\boldsymbol{I}^{GT}_{t} (x)}||_1,}\\
    \mathcal{L}_\mathrm{refine} &= { \sum_{x} ||{\boldsymbol{I}^\mathrm{refine}_{t}(x) }-{\boldsymbol{I}^{GT}_{t}(x)}||_1.}
\end{aligned}
\end{equation}

The total-loss function is defined as:
\begin{equation} \label{eq:totalloss}
\begin{aligned}
    \mathcal{L} &= { {\mathcal{L}_\mathrm{blend} + 2 \cdot \mathcal{L}_\mathrm{refine}}},
\end{aligned}
\end{equation}
where the weight for each loss is empirically determined.
All the networks are trained in an end-to-end manner.

\subsection{Search range analysis in flow estimation}
We can manage the range of receptive fields for flow estimation with the number of pyramid layers. At each pyramid level, the search range is determined with a radius $\boldsymbol{r}$ as mentioned in section~\ref{section:flowupdatemodule}. The totally covered search range is computed by
\begin{equation} \label{eq:search_range}
\begin{aligned}
    {\boldsymbol{R}}_{1} &= \boldsymbol{D}_\text{initial} \cdot \boldsymbol{r},\\
    {\boldsymbol{R}}_{\boldsymbol{L}} &= {\boldsymbol{R}}_{L-1} \cdot \boldsymbol{D}_{L} + \boldsymbol{r},
\end{aligned}
\end{equation}
where the $\boldsymbol{R}_{L}$ means the maximum range of motion coverage, $\boldsymbol{D}_{L}$ is a downsampling ratio according to the pyramid level in the encoder network, and $\boldsymbol{L}$ denotes the total number of pyramid levels.

The proposed model determines the variables to have enough total search range and to have an efficient number of parameters. 
Since we first reduce the image size as a quarter, the downsampling ratio is 2 except for the $\boldsymbol{D}_\text{initial}$. 
The radius $\boldsymbol{r}$ is determined to 4. 
Therefore, the number of pyramid levels should be 6 to cover the 4K dataset, such as X4K1000FPS~\cite{sim2021xvfi}.

\section{Experiments}
We conducted experiments of video frame interpolation for four datasets: X4K1000FPS (4K)~\cite{sim2021xvfi}, VVC (Versatile Video Coding reference dataset, 4K), Vimeo90K (240p)~\cite{xue2019video}, and UCF-101 ($256\times256$)~\cite{soomro2012ucf101}.
\paragraph{X4K1000FPS~\cite{sim2021xvfi}}
The testset in X4K1000FPS contains 15 clips with a length of 33 consecutive frames and each frame is captured by 1000fps. 
We select 8 images with four frames interval in each sequence and conduct the 8 times interpolation to configure a similar test environment in the existing 4K VFI algorithm~\cite{sim2021xvfi}.

\paragraph{VVC} 
VVC contains six 4K videos captured by 30 to 60fps. 
To configure a similar comparison environment with X4K1000FPS, we randomly selected 150 test triplets from VVC with the interval of the input images as 1/30s.

\paragraph{Vimeo90K and UCF-101~\cite{soomro2012ucf101, xue2019video}} Both datasets are standard triplet (240p) sequences to measure the performance of video frame interpolation. 
To validate the performance in the low resolution and low frame-rate videos, we also evaluate for 3,782 triplets in Vimeo90K testset and randomly selected 350 triplets from UCF-101.

\begin{table}[t]
\centering     
\setlength\extrarowheight{0.2em}
\setlength{\tabcolsep}{0.4em}
\begin{tabular}{l?*{2} {c} ?*{2} {c}?{c}}

\specialrule{0.15em}{0.3em}{0.3em}
\multirow{2}{*}{Method}& \multicolumn{2}{c?}{\textbf{X4K1000FPS}} & \multicolumn{2}{c?}{\textbf{VVC}} & \multirow{2}{*}{\#P(M)}\\
& PSNR & SSIM & PSNR & SSIM   \\
\specialrule{0.15em}{0.05em}{0.05em}

$\text{AdaCoF}_{o}$ & 24.35 & 0.7296 & 31.89 & 0.8909 & \multirow{2}{*}{21.8} \\
$\text{AdaCoF}_{f}$ & 25.86 & 0.7691 & 32.67 & 0.8841 &  \\
\specialrule{0.05em}{0.05em}{0.05em}

$\text{CAIN}_{o}$ & 24.50 & 0.7458 & 32.43 & 0.8910 & \multirow{2}{*}{42.8} \\
$\text{CAIN}_{f}$ & 27.36 & 0.8182 & 32.12 & 0.8958 &  \\
\specialrule{0.05em}{0.05em}{0.05em}

$\text{XVFI}_{3}$ & 28.98 & 0.8573 & \textbf{\color{blue}33.17} & \textbf{\color{blue}0.8962} & \multirow{2}{*}{5.5} \\
$\text{XVFI}_{5}$ & \textbf{\color{blue}30.12} & \textbf{\color{blue}0.8707} & 33.08 & 0.8960 & \\

\specialrule{0.15em}{0.3em}{0.3em}
$ \text{Ours}^\text{x4k}_{4} $ & 28.81 & 0.8430 & 33.15 & 0.8977 & \multirow{3}{*}{\textbf{\color{red}4.7}} \\
$ \text{Ours}^\text{x4k}_{5} $ & 30.47 & 0.8719 &  33.19 & 0.8979 \\
$\text{Ours}^\text{x4k}_{6}$ & \textbf{\color{red}{30.51}} & \textbf{\color{red}0.8719} & \textbf{\color{red}33.20} & 
\textbf{\color{red}0.8979} &  \\

\specialrule{0.15em}{0.3em}{0.3em}
\end{tabular}

\caption{Quantitative comparison on X4K1000FPS (4K) and VVC. Note that the numbers in {\color{red} {\textbf{red}}} denote the best performance and numbers in {\color{blue} {\textbf{blue}}} mean the second best performance. Subscript $o$ is official pre-trained weight provided from the author, subscript $f$ denotes fine-tuned model with X4K1000FPS training dataset from the pre-trained weight, and subscripts of Ours and XVFI are the number of pyramid layers for the test. The number of parameters~\#P(M) is expressed in million units.}
\label{tab:XVFIandVVC}
\end{table}

\begin{figure*}
\centering
\includegraphics[width=1.0\linewidth]{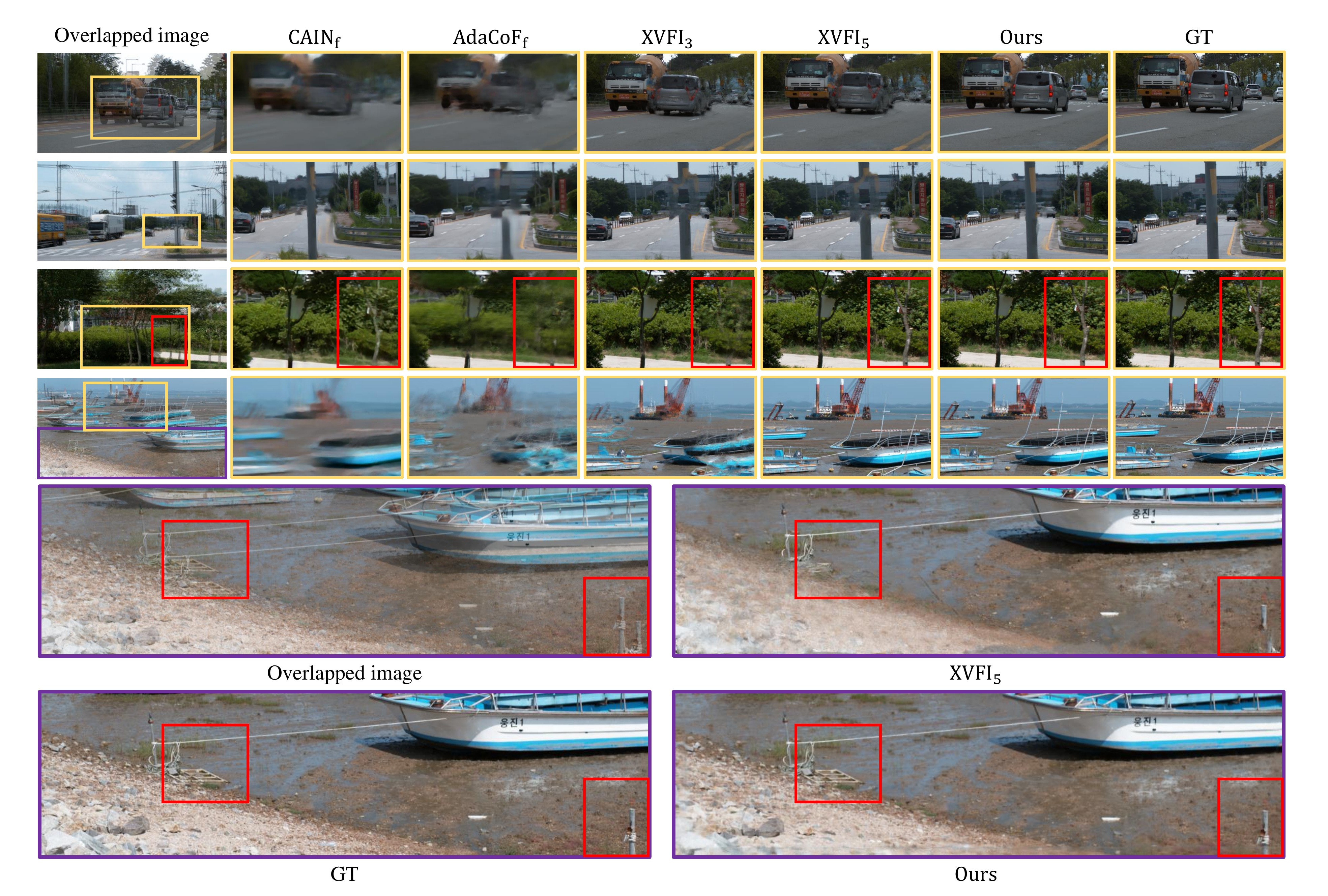}
\caption{Visual comparisons of interpolation results from the X4K1000FPS dataset. Each result is zoomed from the yellow boxed part of the overlapped image. In particular, the purple boxed area is compared for the only two algorithms $\text{XVFI}_\text{5}$ and $\text{Ours}$. Prominent results are highlighted with red boxes.}
\label{fig:Qualitative_Xtest}
\end{figure*}

\subsection{Implementation Details}
We trained our model ($\text{Ours}^\text{x4k}$) for high resolution video using X4K1000FPS~\cite{sim2021xvfi} dataset which contains 4,408 clips. Each clip is composed with a lengths of 65 consecutive frames and image resolution of $768\times768$ cropped from 4K size original image.
During training, two input frames ${\boldsymbol{I}_{0}}$ and ${\boldsymbol{I}_{1}}$, and one target intermediate frame ${\boldsymbol{I}_{t}}$ were grouped for a training sample. 
We designed the training data loader to randomly determine the temporal distance between two inputs as from 2 to 32, and the temporal position of the target frame within the temporal distance of inputs for arbitrary temporal frame interpolation.
The $384\times384$ training image patches were randomly cropped from the original training data. 
We also performed flipping images horizontally and vertically, and swapping the order of frames for data augmentation.

To evaluate the low resolution and low frame-rate dataset, we also trained our another model($\text{Ours}^\text{vimeo}$) using Vimeo90K~\cite{xue2019video} dataset which contains 51,312 triplets of $256\times448$ video frames.
We applied the random cropping as $256 \times 256$, flipping, and swapping on the triplet for the augmentation.
As shown in Equation~\ref{eq:search_range}, we determined the pyramid level $\boldsymbol{L}$ according to the investigated flow magnitude~\cite{sim2021xvfi}. The $\boldsymbol{L}$ for $\text{Ours}^\text{x4k}$ was set to 4, and for $\text{ours}^\text{vimeo}$ was set to 1.

The weights of our model were initialized with the method of He initialization~\cite{he2015delving}. 
We adopted ADAM~\cite{kingma2014adam} optimizer with learning rate $1e-4$, and reduced the learning rate factor of 4 at [100, 150, 180]-th epoch.
We set the batch size as 14 for train $\text{Ours}^\text{x4k}$, and 50 for $\text{Ours}^\text{vimeo}$.
Training takes about two days for $\text{Ours}^\text{x4k}$, and four days for $\text{Ours}^\text{vimeo}$ to fully converge on NVIDIA Tesla V100 GPU.

\begin{table}[t]
\centering     
\setlength\extrarowheight{0.2em}
\setlength{\tabcolsep}{0.4em}
\begin{tabular}{l?*{2} {c} ?*{2} {c}?{c}}
\specialrule{0.15em}{0.3em}{0.3em}
\multirow{2}{*}{Method}& \multicolumn{2}{c?}{\textbf{Vimeo90K}} & \multicolumn{2}{c?}{\textbf{UCF-101}} & \multirow{2}{*}{\#P(M)}\\
& PSNR & SSIM & PSNR & SSIM   \\
\specialrule{0.15em}{0.05em}{0.05em}

AdaCoF  & 34.27 & 0.9714 & 34.95 & 0.9504 & 21.8 \\
CAIN   & 34.65 & 0.9730 & 34.83 & \textbf{\color{red}0.9507} & 42.8 \\
$\text{XVFI}^\text{vimeo}$ & \textbf{\color{red}35.07} & \textbf{\color{red}0.9760} & 34.95 & 0.9505 & 5.5 \\
\specialrule{0.15em}{0.3em}{0.3em}
$\text{Ours}^\text{vimeo}$   & \textbf{\color{blue}34.95} & \textbf{\color{blue}0.9749} & \textbf{\color{red}34.97} & \textbf{\color{blue}0.9506} & \textbf{\color{red}4.7} \\
\specialrule{0.15em}{0.3em}{0.3em}
\end{tabular} 
\caption{Quantitative comparison on Vimeo90K(240p) and UCF-101($256\times256$) with the number of parameters.}
\label{tab:vimeoandUCF101}
\end{table}

\subsection{Quantitative and qualitative analysis}
We conducted the quantitative evaluation for 4K video frame interpolation.
For this experiment, we evaluate the methods on X4K1000FPS testset~\cite{sim2021xvfi} and randomly selected 150 test triplets in VVC. 

We compared our model ($\text{Ours}^\text{x4k}$) with previous video frame interpolation models, CAIN~\cite{choi2020channel}, AdaCoF~\cite{lee2020adacof}, and XVFI~\cite{sim2021xvfi}.
We evaluate both results of methods gathered by using pre-trained weights provided by the author and fine-tuned weight starting from the pre-trained model.
Note that RRPN~\cite{zhang2020flexible} is excluded in the experiment because the official code has not been published yet.

Table~\ref{tab:XVFIandVVC} compares the PSNR and the SSIM measures between our method and other ones.
The results show that our model outperforms the other methods on both 4K datasets with the fewest \# of parameters.
Since the maximum flow magnitude in the X4K1000FPS testset is over 400, our method shows the best accuracy with the pyramid level 6 on the experiment.
The qualitative results described in figure~\ref{fig:Qualitative_Xtest} for X4K1000FPS show that our interpolation results capture the objects which are failed in other ones.  

We also evaluate our model($\text{Ours}^\text{vimeo}$) compared to other previous methods on 3,782 Vimeo90K testset and randomly selected 350 test triplets from UCF-101.

In contrast to the 4K video frame interpolation, the testset in Vimeo90K and UCF-101 contains similar statistics of optical flow magnitude compared to the training environment.
Therefore, we evaluate our model with the pyramid level to 1 which is the same as the level in the training process.

For this experiment, we compare our method and other VFI methods pre-trained with vimeo90K dataset.
Quantitative comparisons of the video frame interpolation results are shown in Table~\ref{tab:vimeoandUCF101}.
These comparisons show our results are on par with the state-of-the-art method in~\cite{sim2021xvfi}, while showing much better results compared to the other video frame interpolation methods.

\subsection{Ablation studies}
\paragraph{Optical flow estimation}
The key contribution of our algorithm is accurate optical flow estimation for high-resolution with fast motion video.
To validate the performance of our flow estimation module, We conduct the ablation study for the model using two existing flow estimation methods, PWC-Net~\cite{sun2018pwc} and RAFT~\cite{teed2020raft}.
Experiments are performed with two conditions for each algorithm. 
One is freezing the flow estimation layers while training the interpolation module, and the other is training the entire network from the pre-trained weights. 
The training is performed under the same conditions in X4K1000FPS dataset~\cite{sim2021xvfi}. 
Table~\ref{tab:ablationstudies} shows the results of ablation studies. 
Since the existing methods are not targeting the large motion and are not familiar with the X4K1000FPS dataset, the fine-tuned model shows improved results compared to each frozen weight model. 
Although the results of fine-tuning have improved, the proposed forward warping based method has better results.
From the results, accurate flow estimation and update modules with large receptive fields are necessary for the 4K VFI.



\paragraph{Refinement module}
To verify the performance of the refinement module, we evaluate our algorithm with and without refinement module for the X4K1000FPS test dataset. 
The optical flow estimation can be failed in high-resolution video with fast object motion.
Furthermore, our interpolation module is designed simply blending two warped frames than other kernel based methods~\cite{bao2019depth, bao2019memc, lee2020adacof, niklaus2017video}, which can cause missing regions in the blended output.
We integrate the dilated convolution layer for the refinement module which is expected to fill the missing parts.
As shown in Table~\ref{tab:ablationstudies}, both PSNR and SSIM are noticeably improved with the refinement module.

\section{Conclusion}
We proposed {ECMNet} to support frame interpolation for high resolution with fast motion video. 
Our model proposed enhanced correlation matching based on bi-directional optical flow estimation from consecutive frames and adopts the advantage of shared weights at each pyramid layer which enables to adjust the number of pyramid levels according to the target video. 
Furthermore, the frame synthesis and refinement networks are designed to synthesize the intermediate frame efficiently.
The proposed network is composed of the integration of convolution layer and differentiable operators which facilitates end-to-end training. 
Our experimental results support the robust performance of the proposed network qualitatively and quantitatively. 
Especially, for the 4K datasets that we targeted, 
{ECMNet} achieve outperformed results against the state-of-the-art algorithms with fewer parameters.

\begin{table}[t]
\centering     
\setlength\extrarowheight{0.2em}
\setlength{\tabcolsep}{0.4em}
\begin{tabular}{l?*{2} {c} ?{c}}

\specialrule{0.15em}{0.3em}{0.3em}
Method & PSNR & SSIM & \#P(M)\\
\specialrule{0.15em}{0.05em}{0.05em}
$\text{PWC-Net}_\text{fr} + \text{synth}$ & 27.39 & 0.8090 & \multirow{2}{*}{11.7} \\
$\text{PWC-Net}_\text{fi} + \text{synth}$ & 28.40 & 0.8183  &  \\
\specialrule{0.05em}{0.05em}{0.05em}
$\text{RAFT}_\text{fr} + \text{synth}$ & 29.17 & 0.8519 &  \multirow{2}{*}{7.7}\\
$\text{RAFT}_\text{fi} + \text{synth}$ & 29.67 & 0.8570 &  \\

\specialrule{0.15em}{0.3em}{0.3em}
$\text{Ours W.O.refine}$ & 30.02 & 0.8609 &  {3.7}\\
$\text{Ours}_\text{full}$ & \textbf{30.51} & \textbf{0.8719} & \textbf{4.7} \\
\specialrule{0.15em}{0.3em}{0.3em}
\end{tabular} 
\caption{Ablation studies for optical flow estimation and refinement network. subscript `fr' denotes freezed pre-trained weight for optical flow estimation network, subscript `fi' is the model fine-tuned the network starting from the pre-trained weight in the training process, and `synth' denotes ours frame synthesis network + refinement network.}
\label{tab:ablationstudies}
\end{table}

{\small
\bibliographystyle{ieee_fullname}
\bibliography{egbib}
}

\end{document}